\pdfoutput=1

\documentclass[11pt]{article}

\usepackage[final]{acl}

\usepackage{times}
\usepackage{latexsym}

\usepackage[T1]{fontenc}

\usepackage[utf8]{inputenc}

\usepackage{microtype}

\usepackage{inconsolata}

\usepackage{graphicx}
\usepackage{inconsolata}
\usepackage{subfigure,graphicx}
\usepackage{booktabs}
\usepackage{multirow}
\usepackage{footnote}
\usepackage{subfigure,graphicx,array}
\usepackage{makecell}
\usepackage{bm}
\usepackage{amsmath}

\usepackage{algorithm}
\usepackage{algorithmic}
\usepackage{caption}
\usepackage{threeparttable}
\usepackage{verbatim}

\newcommand{\bI}{\mathbf{I}}

\newcommand{\bP}{\mathbf{P}}

\newcommand{\bX}{\mathbf{X}}

\newcommand{\bh}{\mathbf{h}}

\newcommand{\bx}{\mathbf{x}}

\newcommand{\ours}{\texttt{PPlug}}
\newcolumntype{P}[1]{>{\centering\arraybackslash}p{#1}}

\title{LLMs + Persona-Plug = Personalized LLMs}

\author{Jiongnan Liu$^{1}$, Yutao Zhu$^{1}$, Shuting Wang$^{1}$, Xiaochi Wei$^{3}$ \\ \textbf{Erxue Min$^{3}$, Yu Lu$^{3}$, Shuaiqiang Wang$^{3}$, Dawei Yin$^{3}$, Zhicheng Dou$^{1,2}$} \\
  $^{1}$Gaoling School of Artificial Intelligence, Renmin University of China \\  
  $^{2}$Engineering Research Center of Next-Generation Intelligent Search and Recommendation, MOE\\
  $^{3}$Baidu Inc.\\
  \texttt{liujn@ruc.edu.cn, yutaozhu94@gmail.com, dou@ruc.edu.cn}}

\begin{document}
\maketitle

\begin{abstract}
Personalization plays a critical role in numerous language tasks and applications, since users with the same requirements may prefer diverse outputs based on their interests. This has led to the development of various personalized approaches aimed at adapting large language models (LLMs) to generate customized outputs aligned with user preferences.
Some of them involve fine-tuning a unique personalized LLM for each user, which is too expensive for widespread application. Alternative approaches introduce personalization information in a plug-and-play manner by retrieving the user's relevant historical texts as demonstrations. 
However, this retrieval-based strategy may break the continuity of the user history and fail to capture the user's overall styles and patterns, hence leading to sub-optimal performance. 
To address these challenges, we propose a novel personalized LLM model, \ours{}. It constructs a user-specific embedding for each individual by modeling all her historical contexts through a lightweight plug-in user embedder module.
By attaching this embedding to the task input, LLMs can better understand and capture user habits and preferences, thereby producing more personalized outputs without tuning their parameters. Extensive experiments on various tasks in the language model personalization (LaMP) benchmark demonstrate that the proposed model significantly outperforms existing personalized LLM approaches.

\end{abstract}

\section{Introduction}
\label{sec:intro}
Large language models (LLMs) have demonstrated extraordinary capabilities in natural language understanding, generation, and reasoning~\cite{llm_survey,LLM1,llm4ir,reta-llm,inters}, becoming increasingly essential tools for assisting with everyday tasks. However, the dominant usage pattern of LLMs follows a \textit{one-size-fits-all} approach, where similar responses are provided to different users given the same input. While sampling-based decoding strategies can introduce some diversity, this approach fails to account for individual user preferences, reducing engagement in human-machine interactions. This problem is even severe in scenarios requiring tailored responses to align with subjective user profiles, such as drafting personalized speeches. Consequently, personalized LLMs have attracted significant interest in both industry and academic research~\cite{LAMP, hydra, longlamp, OPPU, llm_per_survey1, llm_per_survey2}.

\begin{figure*}
    \centering
    \includegraphics[width=1.0\linewidth]{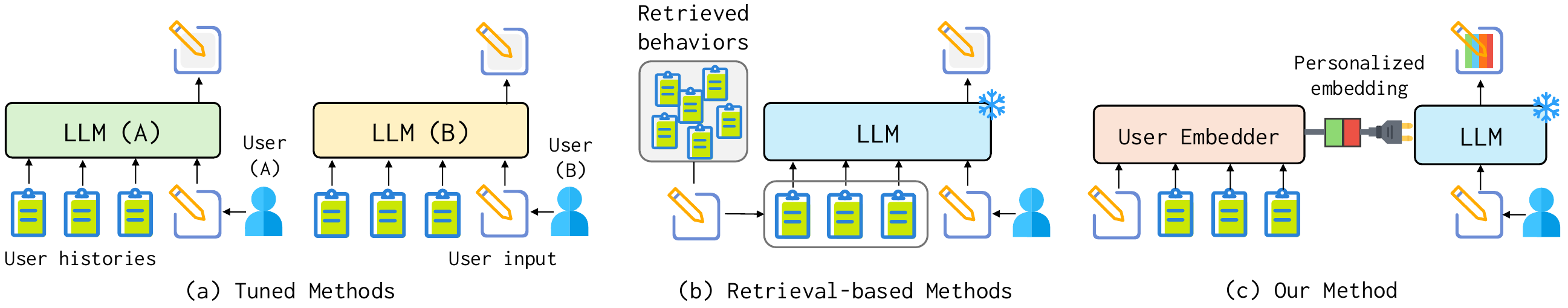}
    \caption{The comparison of our proposed personalized LLM and previous personalized LLM approaches}
    \label{fig:intro}
\end{figure*}

A straightforward strategy for building personalized LLMs is to fine-tune a specific LLM on individual user data, allowing the model to learn the specific patterns and preferences of each user~\cite{OPPU, OPPU_2, MiLP}. While effective, this method requires extensive computing resources for both training and inference, making it challenging for deployment in real applications. These approaches also suffer when users only have limited training data~\cite{lamp_compare}.
Another way to achieve personalization is directly feeding all user histories into the LLM, and then generating tailored results according to the current user requests~\cite{longperllm}. This strategy avoids the need for additional model training but is often constrained by the maximum input length of the LLM, resulting in unsatisfying performance. To tackle this problem, recent studies have proposed leveraging retrieval models to select relevant behaviors from user histories based on the user input~\cite{longlamp, LAMP, RSPG, PAG}. These retrieved behaviors are then used as in-context demonstrations to guide the LLM in generating personalized outputs. 

While this strategy can introduce some degree of personalization, it is not always reliable. For producing personalized results, it is more important for LLMs to understand the users' overall styles than to refer to specific histories.  Unfortunately, the retrieval process typically focuses on relevance to the current input rather than identifying deeper user preferences embedded in all historical data. As a result, the selective utilization of user histories in retrieval-based personalized LLMs may disturb the model to capture user comprehensive manners and lead to sub-optimal performance.

Therefore, a better strategy for personalizing LLMs is to plug the user's holistic styles into the LLMs without modification of their structures and parameters. 
To achieve this, we propose a persona-plug (\ours{}) model. It involves a lightweight plug-in user embedder module that embeds user historical patterns into a single user-specific embedding in input for LLMs to refer to. 
In the user embedder module, we first develop a user behavior encoder to represent a user's each historical behavior into a dense vector. Then, an input-aware personal aggregator synthesizes all these vectors into a user-specific personal embedding according to their relevance to current task inputs. 
This unique personal embedding from all histories is supposed to capture the user's general patterns in language tasks. After obtaining this personalized embedding, we directly attach it to the current input to guide fixed LLMs in tailoring their outputs according to user preferences. In this way, our \ours{} model can better perform personalized 
tasks relying on the extracted user's comprehensive personal patterns in a plug-and-play strategy, shown in Figure~\ref{fig:intro}. Furthermore, \ours{} model can also be optimized in an end-to-end manner 
by all users' data, which is more efficient and effective compared with resorting to the limited data of each user to fine-tune personalized LLMs.

Experiments on six tasks in the public language model personalization (LaMP) benchmark~\cite{LAMP} show that our proposed \ours{} model achieves significant improvements over existing personalized LLM models from 1.4\% to 35.8\%. The main contribution of our work is three-fold:

(1) To better guide LLMs in personalized language generation, we propose a novel personalization framework that only attaches one user's personal embedding for LLMs to refer to.

(2) Compared with tuning a specific LLM for each user, the proposed \ours{} model follows the plug-and-play paradigm and brings no additional parameters to LLMs.

(3) Compared with retrieval-based LLMs, \ours{} can capture user holistic patterns and preferences, leading to better personalization performance.

\section{Related Work}
With the rising development of large language model techniques in many NLP applications and tasks~\cite{llm_survey, llm4ir}, personalization in LLMs has attracted attention, and many approaches have been recently proposed~\cite{LAMP, RSPG, hydra, longlamp, OPPU, OPPU_2, PAG, perllm}. These approaches facilitate LLMs with the personal content of users to generate customized outputs. Most of them can be categorized into the following two kinds: 

\noindent\textbf{Fine-tuned Personalized LLMs.}\quad The simple strategy for personalized language generation is to tune a unique LLM for each user based on their own data. However, fine-tuning all parameters in LLMs is too expensive; approaches in this category mainly devise the parameter-efficient fine-tuning (PEFT) technique to tune LLMs. Specifically, OPPU~\cite{OPPU} adopts the LoRA methods~\cite{lora} to tune the Llama model~\cite{llama} for each user. ~\citet{OPPU_2} further improve it by clustering users into different groups and tuning a model for each group. ~\citet{hydra} optimize a distinct language head for each user to tailor LLMs output. ~\citet{MiLP} modify the model by searching for the best configuration of PEFT methods for each user.

\noindent\textbf{Retrieval-based Personalized LLMs.}\quad The fine-tuned personalized LLMs need to train the LLMs for each user separately, which introduces huge computation costs and is difficult to devise in real applications. Retrieval-based personalized LLMs leverage personalized information from another perspective without tuning LLMs. Inspired by the success of the retrieval-augmented generation (RAG) strategy in question-answering tasks~\cite{spring}, these approaches retrieve relevant documents from user histories as in-context demonstrations for LLMs to produce personalized texts. ~\citet{LAMP} explore these methods by applying different retrieval models. ~\citet{RSPG} further improve it by optimizing the retrieval model through rewards calculated based on the LLM-generated outputs. They also explore the selection of different retrieval methods while facing different inputs. There also exist models that directly utilize all user histories to prompt LLMs or apply language models to generate text-based summaries as prompts~\cite{PAG, longperllm, PAG2}. However, these approaches cannot handle user histories that are extremely long due to the input length limits.

Some approaches~\cite{rec1,rec2,rec3} in recommendation areas also aim at personalizing language models. However, they only adopt small language models such as T5-base model and the private PaLM 2-XXS~\cite{palm} model and mainly focus on recommendations instead of language tasks requiring extensive world knowledge.

\section{Methodology}

\begin{figure*}[!tbp]
  \centering
  \includegraphics[width=0.88\linewidth]{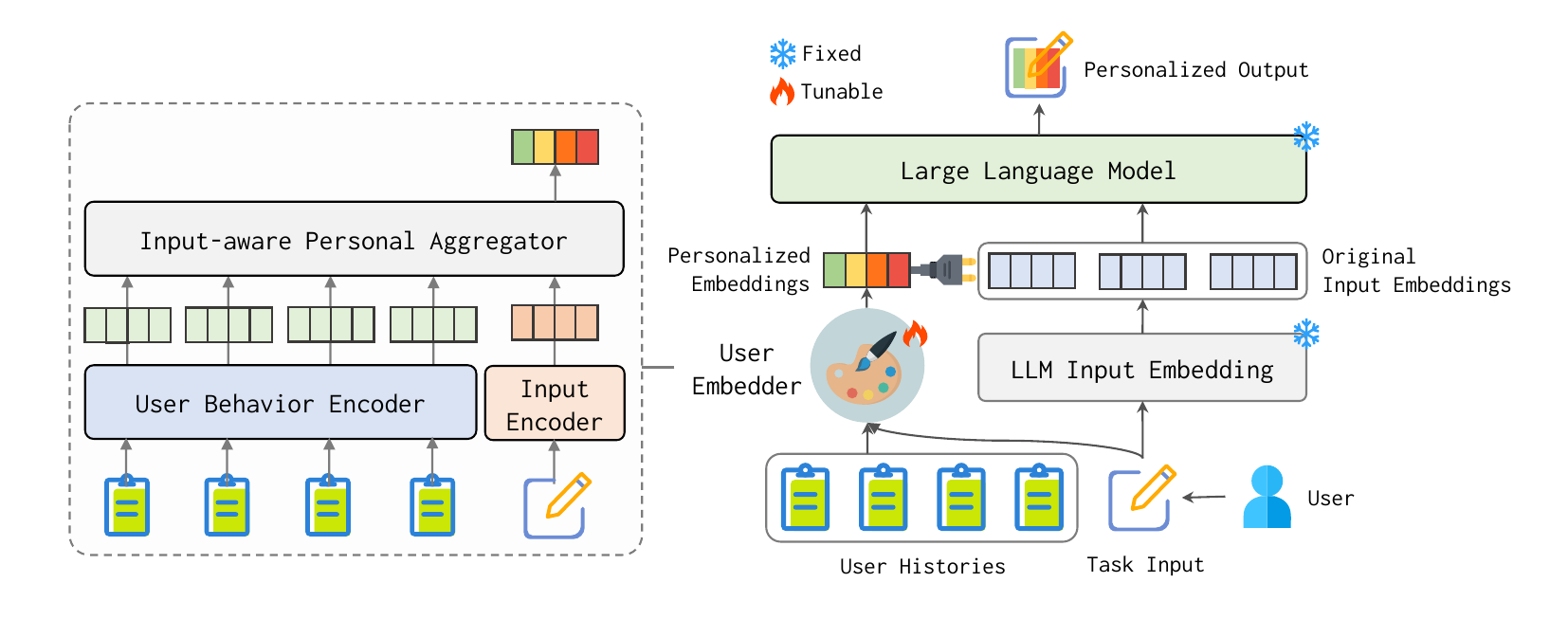}
  \caption{The overall framework of the proposed \ours{} model.}
  \label{fig:framework}
\end{figure*}

Personalized large language models (LLMs) aim to satisfy users' specific demands and preferences by tailoring responses based on users' historical behaviors.\footnote{Other personal information, such as user attributes, can also be used for personalization. However, due to the absence of such data in the current dataset, we follow existing studies~\cite{RSPG,LAMP} and focus solely on users' histories.} Following existing studies~\cite{RSPG,LAMP}, the personalization task can be defined as: for a certain user $u$, generating a personalized response $y^u$ to a given user input $x^u$, utilizing the user's historical behaviors $H^u = [h^u_1,\cdots,h^u_n]$. Each user behavior $h^u_i$ corresponds to historical interactions similar in nature to the current input $x^u$. For example, if a user requests assistance with generating a title for a research paper, their historical behaviors may include titles and abstracts they have previously created for papers.

In this work, we introduce a method for LLM personalization called the persona-plug (\ours{}) with a plug-in user embedder module. It encodes each historical behavior of a user into a dense vector and aggregates these embeddings into a single personal embedding considering the current input $x^u$. This personal embedding is then incorporated into the input to guide a fixed LLM in generating personalized responses. \ours{} is a lightweight, plug-and-play approach, where each user has a distinct personal embedding calculated by the shared user embedder. The  LLM uses these embeddings as input without requiring any additional modification to its own parameters. An overview of our proposed \ours{} method is shown in Figure~\ref{fig:framework}.

\subsection{User Behavior Encoder}
User behaviors often reflect how a user deals with a specific task, which contains valuable personal preferences and linguistic patterns. Therefore, effectively representing user behaviors is a critical step for personalization. Inspired by recent studies on sentence embedding and dense retrieval~\cite{simcse, contriever}, we employ a user behavior encoder to obtain user behavior representations. Specifically, for each user historical behavior $h^u_i$, we leverage an encoder-based model ${\textrm{Enc}^\textrm{his}}(\cdot)$ to encode $h^u_i$ as a vector $\bh^u_i$:
\begin{equation}
    \bh^u_i = {\textrm{Enc}^\textrm{his}}(h^u_i).
\end{equation}
Similarly, the representation of the current user input $x^u$ is computed as:
\begin{equation}
\bx^u={\textrm{Enc}^\textrm{input}}(x^u), 
\end{equation}
where the ${\textrm{Enc}^\textrm{input}}(\cdot)$ denotes the encoder specific to the user’s current input, such as personalized product review.\footnote{In our implementation, we use the BGE-base-en-v1.5 model~\cite{bge_1, bge_2} as the encoder, \url{https://huggingface.co/BAAI/bge-base-en-v1.5}.} All tasks are introduced in Section~\ref{sec:task}. To ensure efficient training of our proposed model, we freeze the parameters of ${\textrm{Enc}^\textrm{his}}$ and only fine-tune the input encoder ${\textrm{Enc}^\textrm{input}}$.

We choose small-sized encoder-based models for two primary reasons: (1) Bi-directional attention can effectively capture interactions across all tokens in user behaviors. Previous studies in information retrieval have demonstrated that encoder models can effectively condense document information into compact and dense representations~\cite{ir_encoder_1}. (2) A lightweight encoder improves the efficiency of both optimization and inference in our \ours{} model. In our implementation, the encoder model introduces approximately $220$M parameters, accounting for only $3.1\%$ of the total parameters in a 7B LLM.

\subsection{Input-aware Personal Aggregator}
\label{subsec:per_aggr}
After obtaining representations of a user's historical behaviors and input, the next step is to aggregate them into a comprehensive personal embedding. A common approach is to treat each historical behavior as equally important and simply average them to represent the user profile. However, previous studies on personalized search and recommendation~\cite{HRNN, sasrec, explips} show that the importance of historical behaviors for the ongoing task should consider their relevance to the current input. For example, in generating academic titles based on abstracts, the model would benefit from prioritizing historical titles and abstracts that align more closely with the topics of the current abstract. Therefore, to improve task performance, historical behaviors that are more relevant to the current input should be assigned higher weights. To this end, we devise an attention mechanism that dynamically assigns weights to each historical behavior based on its relevance to the current user input. The personal embedding is calculated as follows:
\begin{align}
    w_i &= \frac{\exp ({\bx^u}^{\top} \bh^u_i)}{\sum_k \exp ({\bx^u}^{\top} \bh^u_k)}, \\
    \bP^u &= \sum_i w_i \cdot  \textrm{Proj}(\bh^u_i), \label{eq:per_emb}
\end{align}
where $\textrm{Proj}(\cdot)$ projects the user embeddings from the encoder space to the LLM representation space by a 2-layer MLP, and $\bP^u$ denotes the calculated personal embedding. In this manner, \ours{} can mimic the retrieval manipulation in the retrieval-based strategy to make LLMs pay more attention to historically relevant behaviors. However, different from retrieval-based personalization methods that focus only on the most relevant histories, our approach integrates all user behaviors. This enables the personal embedding $\bP^u$ to capture a more holistic representation of the user’s preferences and patterns, enhancing the \ours{}'s ability to tailor personalized outputs.

\subsection{\ours{} for LLM Personalization}
\label{subsec:methodall}
Once the personal embedding $\bP^u$ is obtained, it is attached to the input to guide a fixed LLM in generating personalized outputs. Specifically, given the user's current input $x^u$ and previously generated personalized content $y^{u}_{<i}$, the next token prediction loss is defined as:
\begin{align}
\bX^u_i &= [\bI;\bP^u;\textrm{Emb}_{\textrm{LLM}}(x^u);\textrm{Emb}_{\textrm{LLM}}(y^u_{<i})], \label{eq:loss}\\
\mathcal{L} &= - \sum_u \sum_i \log p_{\textrm{LLM}} (y^u_i |\bX^u_i), 
\end{align}
where $\textrm{Emb}_{\textrm{LLM}}(\cdot)$ denotes the LLM's embedding layer, $p_{\textrm{LLM}}$ is the predicted token distributions. Note that in addition to incorporating a personal embedding, we introduce a trainable instruction embedding $\bI$ into the input. This is inspired by several recent studies on instruction tuning~\cite{instruction_1,bge_1}, which have shown that including an instruction embedding helps the LLM better understand and perform the task. Particularly, the LLM used in the \ours{} model is fixed and only the instruction embedding $\bI$, the input encoder ${\textrm{Enc}^\textrm{input}}(\cdot)$, and the projector $\textrm{Proj}(\cdot)$ are tuned, which is efficient for application.

\subsection{Comparison with Previous Models}
To highlight the advantages of the \ours{} model, we provide a comparison with existing methods.

\noindent\textbf{\ours{} vs. Fine-tuned Methods}\quad 
Both the fine-tuned personalization methods~\cite{MiLP, OPPU} and our \ours{} model train the personalized framework to capture user general interests to guide personalized language generation, leading to promising performances. Besides, \ours{} model has two additional advantages: 
(1)~In training, unlike fine-tuned methods that require training a separate LLM for each user relying on their own limited data, \ours{} trains a shared encoder to capture personalized user information using all data more efficiently and effectively. 
(2)~In inference, \ours{} operates in a plug-and-play manner, where a single LLM is used for all users, with user-specific personalized embeddings provided as input. This is highly advantageous for LLM service providers, as it enables the deployment of a single model to deliver effective personalization across users, streamlining infrastructure and maintenance.

\noindent\textbf{\ours{} vs. Retrieval-based Methods}\quad 
Retrieval-based methods achieve personalization by selecting relevant user historical behaviors. Similarly, \ours{} incorporates an input-aware attention mechanism to evaluate the relevance of each behavior. However, unlike retrieval-based approaches that focus only on the most relevant behaviors, \ours{} assigns dynamic weights to all behaviors. This allows it to capture a more comprehensive view of the user's general preferences across their entire history, leading to improved personalization outputs. 
Furthermore,  while retrieval-based LLMs need to record the whole user histories for retrieval, \ours{} only utilizes user embeddings, which can be produced by users themselves through the lightweight user embedder, which can better protect user privacy.

\section{Experiments}
\subsection{Datasets and Metrics}\label{sec:task}
\paragraph{Datasets} We conduct experiments using the public Language Model Personalization (LaMP) benchmark~\cite{LAMP}, which consists of seven different personalization tasks. Consistent with previous studies~\cite{OPPU, hydra, PAG, OPPU_2}, we evaluate model performance on six tasks, excluding the Personalized Email Subject Generation task (LaMP-6), as it is not publicly available. Concretely, the six tasks include three personalized text classification tasks:
(1) LaMP-1 Personalized Citation Identification; (2) LaMP-2 Personalized Movie Tagging; (3) LaMP-3 Personalized Product Rating, and three personalized text generation tasks: (4) LaMP-4 Personalized News Headline Generation; (5) LaMP-5 Personalized Scholarly Title Generation; and (6) LaMP-7 Personalized Tweet Paraphrasing. 
We use the time-based datasets provided by the LaMP benchmark, in which the data for each user is split into train, validation, and test sets in chronological order. 
Detailed information can be found in Appendix~\ref{apd:dataset}.

\paragraph{Evaluation Metrics} We use the default metrics in LaMP benchmark to evaluate the performance of each task: accuracy for LaMP-1, accuracy and F1-measure for LaMP-2, mean absolute error (MAE) and root mean squared error (RMSE) for LaMP-3, and ROUGE-1 and ROUGE-L~\cite{rouge} for LaMP-4, LaMP-5, and LaMP-7. For MAE and RMSE, lower values indicate better performance, as these metrics measure the discrepancy between predictions and ground-truth. For all other metrics, higher values correspond to better performance.

\begin{table*}[!ht]
 \center
 \small
 \caption{Performance of all models on six LaMP tasks. ``Valid'' and ``Test'' refer to the results on the validation and test sets, respectively. The best results are in \textbf{bold}.}\label{tlb:overall}
  \setlength{\tabcolsep}{1.1mm}\resizebox{0.99\linewidth}{!}{{
  \begin{tabular}{lccccccccccc}
  	\toprule
        \multirow{2}{*}[-2pt]{Dataset} & \multirow{2}{*}[-2pt]{Metric} & Ad-hoc & \multirow{2}{*}[-2pt]{FTP} & \multicolumn{3}{c}{Naive RBP} & \multicolumn{4}{c}{Optimized RBP} & \multirow{2}{*}[-2pt]{\ours{}} \\
        \cmidrule(lr){3-3} \cmidrule(lr){5-7} \cmidrule(lr){8-11} 
        & & FlanT5-XXL&  & BM25 & Recency & Contriever & \texttt{ROPG-RL} & \texttt{ROPG-KD} & \texttt{RSPG-Pre} & \texttt{RSPG-Post} & \\
        \midrule
        \multirow{2}{*}{LaMP-1} & Valid Accuracy $\uparrow$ &0.498 & - & 0.629 & 0.639 & 0.641 &\textbf{0.682} &0.676 &0.672 &0.670 &{0.680} \\
        & Test Accuracy $\uparrow$ &0.502 & 0.506 & 0.626 & 0.622 & 0.636 &0.655 &0.668 &0.663 &0.672 & \textbf{0.700} \\
        \midrule
        \multirow{4}{*}{LaMP-2} & Valid Accuracy $\uparrow$ &0.326 & -& 0.345 & 0.361 & 0.362 & 0.365 & 0.365 & 0.391 & 0.416 & \textbf{0.565}\\
         & Valid F1 $\uparrow$ &0.255 & - & 0.282 &0.291 & 0.282 & 0.292 & 0.291 & 0.312  & 0.337 & \textbf{0.501} \\
        & Test Accuracy $\uparrow$ &0.359 & 0.360 & 0.387 & 0.377 & 0.396 & 0.391 & 0.396 & 0.405 & 0.430 & \textbf{0.559}\\
         & Test F1 $\uparrow$ &0.276 &0.278 & 0.306 &0.295 & 0.304 & 0.300 & 0.306 & 0.314  & 0.339 & \textbf{0.495} \\
        \midrule
        \multirow{4}{*}{LaMP-3} & Valid MAE $\downarrow$ &0.335 & - & 0.293 &0.305 & 0.297 & 0.273 & 0.274 & 0.266 & 0.246 & \textbf{0.231} \\
         & Valid RMSE $\downarrow$ &0.639 & - &0.585 &0.596 & 0.592 & 0.561 & 0.566 & 0.560 & 0.539 & \textbf{0.534} \\
        & Test MAE $\downarrow$ &0.308 & 0.301 & 0.298 &0.296 & 0.299 & 0.286 & 0.290 & 0.282 & 0.264 & \textbf{0.242} \\
         & Test RMSE $\downarrow$ &0.611 & 0.600 &0.611 &0.605 & 0.616 & 0.591 & 0.604 & 0.585 & 0.568 & \textbf{0.557} \\
        \midrule
        \multirow{4}{*}{LaMP-4} & Valid ROUGE-1 $\uparrow$ &0.173 & - & 0.192 & 0.194 & 0.190 & 0.190 & 0.193 & 0.195 & 0.207 & \textbf{0.216}\\
         & Valid ROUGE-L $\uparrow$ &0.157 & - & 0.175 & 0.177 & 0.174 & 0.174 & 0.176 & 0.179 & 0.188 & \textbf{0.197} \\
        & Test ROUGE-1 $\uparrow$ &0.176 & 0.178 & 0.186 & 0.189 & 0.183 & 0.191 & 0.187 & 0.190 & 0.203 & \textbf{0.211}\\
         & Test ROUGE-L $\uparrow$ &0.160 & 0.163 & 0.171 & 0.173 & 0.169 & 0.177 & 0.172 & 0.176 & 0.186 & \textbf{0.193} \\

        \midrule
        \multirow{4}{*}{LaMP-5} & Valid ROUGE-1 $\uparrow$ &0.472 & - & 0.467 & 0.469 & 0.471 & 0.473 & 0.472 & 0.479 & 0.480 & \textbf{0.487}\\
         & Valid ROUGE-L $\uparrow$ &0.419 & - & 0.419 & 0.422 & 0.421 & 0.425 & 0.423 & 0.429 & 0.429 & \textbf{0.436} \\
        & Test ROUGE-1 $\uparrow$ &0.478 & 0.478 & 0.477 & 0.475 & 0.483 & 0.475 & 0.477 & 0.483 & 0.480 & \textbf{0.487}\\
         & Test ROUGE-L $\uparrow$ &0.428 & 0.429 & 0.427 & 0.426 & 0.433 & 0.427 & 0.428 & 0.431 & 0.429 & \textbf{0.439} \\

        \midrule
        \multirow{4}{*}{LaMP-7} & Valid ROUGE-1 $\uparrow$ &0.454 & - &0.451 &0.452 & 0.440 & 0.458 & 0.451 & 0.460 & 0.468 & \textbf{0.534} \\
         & Valid ROUGE-L $\uparrow$ &0.401 & - & 0.401 &0.402 & 0.391 & 0.407 & 0.402 & 0.409 & 0.416 & \textbf{0.484} \\         
        & Test ROUGE-1 $\uparrow$ &0.449 & 0.449 &0.446 &0.444 & 0.440 & 0.448 & 0.441 & 0.450 & 0.461 & \textbf{0.537} \\
         & Test ROUGE-L $\uparrow$ &0.396 & 0.397 & 0.394 &0.393 & 0.390 & 0.397 & 0.391 & 0.400 & 0.409 & \textbf{0.484} \\       
	\bottomrule
  \end{tabular}}
  }
\end{table*}

\subsection{Implementaion Details} 
We use FlanT5-XXL (11B)~\cite{flant5} as the default LLM, which is consistent with previous studies~\cite{LAMP, RSPG}.
We use BGE-base-en-v1.5~\cite{bge_2} as our default history and input encoder. Experimental results on different LLMs and encoder models are provided in Section~\ref{subsec:llmandencder}. The maximum input lengths are set to 256 tokens for the LLMs and 512 tokens for the encoder. We employ beam search~\cite{beamsearch} with a beam size of 4 during generation. 
We train our \ours{} model for 2 epochs across all tasks, except for LaMP-3, where 1 epoch is sufficient due to the larger dataset size. 
The batch size in all experiments is set to 64. 
The codes are available in \url{https://github.com/rucliujn/PPlug}.

\subsection{Baselines} 
We compare our \ours{} model with the following baselines covering four kinds of approaches:

(1)~\textbf{Ad-hoc methods}: We use FlanT5-XXL to generate outputs solely based on the original task inputs. It serves as a non-personalized baseline.

(2)~\textbf{Fine-tuned Personalization methods (FTP)}: Fine-tuning a specific LLM for each user requires extensive computational resources for training and inference (10,000 hours of A100 GPU computation and 18 TB checkpoint storage are needed for LaMP benchmark as reported in \citet{lamp_compare}, while \ours{} model only requires about 100 hours and 150GB storage), making it difficult to reproduce. Therefore, we directly copy the results of PEFT personalization reported in ~\cite{lamp_compare} which only contains the test results without validation ones under the same evaluation setup as \ours{}. They applied LoRA on FlanT5-XXL to tune a specific LLM for each user, similar to OPPU~\cite{OPPU}.

(3)~\textbf{Naive retrieval-based personalization methods (Naive RBP)}: We employ BM25~\cite{bm25}, Recency, and Contriever~\cite{contriever} methods to retrieve the top-4 user historical behaviors as demonstrations for FlanT5-XXL to produce personalized outputs. These methods are not tuned for personalization tasks and thus are referred to as naive RBP.

(4)~\textbf{Optimized Retrieval-based Personalization (Optimized RBP)}: \texttt{ROPG-RL}, \texttt{ROPG-KD}, \texttt{RSPG-Pre}, and \texttt{RSPG-Post} are four baseline methods designed by \citet{RSPG}. \texttt{ROPG-RL} and \texttt{ROPG-KD} optimize the Contriever-based retrieval model by reinforcement learning and knowledge distillation strategies according to the evaluation metrics. \texttt{RSPG-Pre} and \texttt{RSPG-Post} introduce a retrieval selection module that selects the optimal retrieval model from multiple candidates based on task inputs and model outputs, respectively.

\subsection{Experimental Results}
\label{subsec:exp}
\begin{table*}[!ht]
 \centering
 \small
 \caption{Overall performance of models with different LLMs and encoders on the validation set. We use Acc to abbreviate Accuracy and R to abbreviate ROUGE respectively.
 }\label{tlb:llmandencoder}
   \setlength{\tabcolsep}{1.2mm}{
  \begin{tabular}{llcccccccccccc}
  	\toprule
        \multirow{2}{*}[-2pt]{LLM} & \multirow{2}{*}[-2pt]{Encoder} & LaMP-1 & \multicolumn{2}{c}{LaMP-2} & \multicolumn{2}{c}{LaMP-3} & \multicolumn{2}{c}{LaMP-4} & \multicolumn{2}{c}{LaMP-5} & \multicolumn{2}{c}{LaMP-7} & \multirow{2}{*}[-2pt]{\# Best} \\
        \cmidrule(lr){3-3} \cmidrule(lr){4-5} \cmidrule(lr){6-7} \cmidrule(lr){8-9} \cmidrule(lr){10-11} \cmidrule(lr){12-13}
        & & Acc $\uparrow$ &  Acc $\uparrow$ &  F1 $\uparrow$ &  MAE $\downarrow$ &  RMSE $\downarrow$  & R-1 $\uparrow$ & R-L $\uparrow$ & R-1 $\uparrow$ & R-L $\uparrow$ & R-1 $\uparrow$ & R-L $\uparrow$ \\
        \midrule
        FlanT5-XL & BGE & 0.636 & 0.463 & 0.375 & 0.242 & 0.537 & 0.193 & 0.174 & 0.478 & 0.424 & 0.509 & 0.456 & 0 \\
        FlanT5-XXL & BGE & 0.680 & 0.565 & 0.501 & \textbf{0.231} & 0.534 & \textbf{0.216} & \textbf{0.197} & \textbf{0.487} & \textbf{0.436} & \textbf{0.534} & \textbf{0.484} & 7 \\
        FlanT5-XXL & Contriver& \textbf{0.687} & 0.553 & 0.501 & 0.236 & \textbf{0.527} & \textbf{0.216} & \textbf{0.197} & 0.485 & \textbf{0.436} & 0.535 & 0.482 & 5 \\
        Llama 2 7B & BGE& 0.663 & 0.585 & 0.540 & 0.259 & 0.581 & 0.212 & 0.194 & 0.467 & 0.418 & 0.503 & 0.450 & 0\\
        Llama 2 7B & Contriver & 0.611 & \textbf{0.589} & \textbf{0.547} & 0.261 & 0.582 & \textbf{0.216} & 0.196 & 0.466 & 0.417 & 0.504 & 0.450 & 3 \\
        
	\bottomrule
  \end{tabular}
  }
\end{table*}

The results on the validation and test sets are shown in Table~\ref{tlb:overall}. Generally, \ours{} achieves the best performance, demonstrating its superiority on personalization tasks. Furthermore, we observe that:

(1) Fine-tuned Personalization methods (FTP) only achieve subtle improvements over non-personalized methods (ad-hoc). The reason may be that fine-tuning personalized LLMs requires sufficient user behavior data, but most users in the LaMP benchmark only have limited histories~\cite{lamp_compare}. 
(2)~Both retrieval-based methods (RBP) and \ours{} can achieve better performance. This indicates that incorporating user historical behaviors is an effective way to capture user personal preferences. 
(3)~Compared to naive RBP, optimized RBP can perform better. This is consistent with our speculation, as the retrievers in naive RBP are not optimized for personalized generation tasks, and tuning the retrievers with the feedback from LLMs' output is beneficial for personalization tasks. 
(4)~Our \ours{} outperforms all baselines in almost all tasks. Specifically, the relative improvements of \ours{} over the best baseline (\texttt{RSPG-Post}) are from 1.4\% to 35.8\%. These improvements confirm that our idea of comprising user historical behaviors into a single personal representation and facilitating LLMs to perform personalized tasks is very effective. (5) Compared with FTP methods, the \ours{} model can utilize data from all users instead of each user's own limited data, which is more effective. 
Compared with \texttt{ROPG} and \texttt{RSPG} that leverage reinforcement learning and knowledge distillation techniques to optimize models, our \ours{} can be directly optimized in an end-to-end manner, which is much more efficient.

\begin{figure}
    \centering
    \includegraphics[width=.85\linewidth]{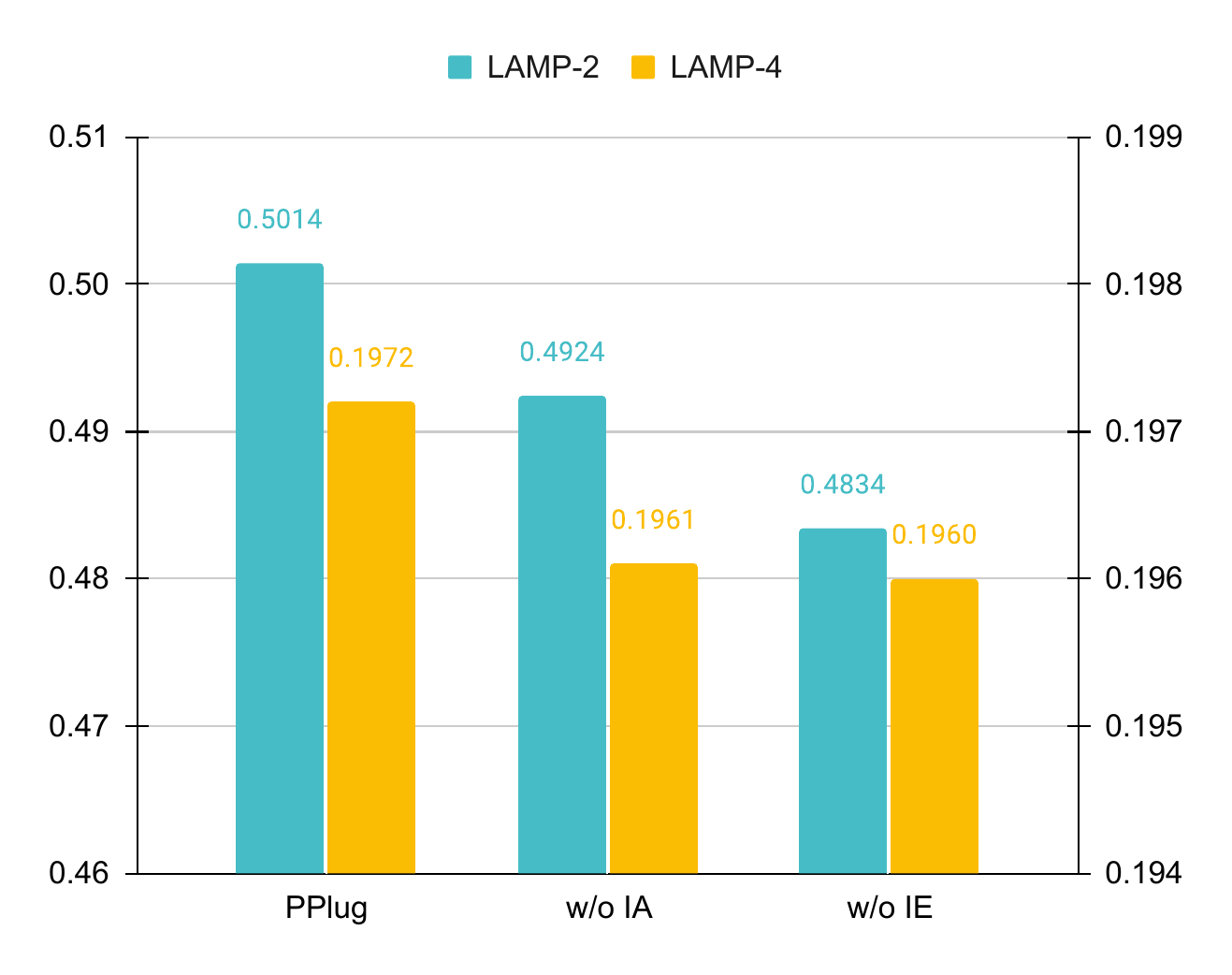}
    \caption{Overall performance of ablation models on the validation set.}
    \label{fig:abl}
\end{figure}

\subsection{Further Analysis}
We further conduct a series of experiments to analyze our \ours{} method. Since the test data are held out by the benchmark organizers, we mainly report the results on the validation set in this section. 

\paragraph{LLM and Encoder Analysis}
\label{subsec:llmandencder}
By default, we use FlanT5-XXL and BGE-base as the LLM and encoder. To investigate their impact on the final performance, we conduct more experiments by replacing them with other models. Specifically,  we replace the LLM by FlanT5-XL~\cite{flant5} and Llama 2 7B Chat~\cite{llama2} and the encoder model by Contriever~\cite{contriever} and test the performance of these variants. The results are shown in Table~\ref{tlb:llmandencoder}. We can find: (1)~When using FlanT5-XXL as the backbone LLM, \ours{} with both BGE-base and Contriever can achieve comparable performance, and both of them can outperform previous personalization methods significantly. This result clearly validates the robustness of our method. (2)~When using the BGE-base encoder, \ours{}'s performance is positively correlated with the size of the LLM (FlanT5-XXL 11B > Llama 2 7B > FlanT5-XL 3B). This result is consistent with the scaling law~\cite{scaling_law}, where larger models have stronger capabilities and perform better on NLP tasks. 


\paragraph{Ablation Study}
\label{para:abl}
In our proposed \ours{}, we design an input-aware personal aggregator that dynamically constructs personal embeddings based on the current task input  (Section~\ref{subsec:per_aggr}). Additionally, we employ an instruction embedding to capture global patterns relevant to specific tasks (Section~\ref{subsec:methodall}). To investigate the effect of these components, we perform an ablation study. Due to limited space, we report the results on two representative tasks: LaMP-2 and
LaMP-4 
tasks. 
The performance observed in these tasks is consistent with trends across other tasks. 
Complete results can be found in Table~\ref{fig:abl} in Appendix~\ref{apd:ablexp}. 

(1) \textbf{Impact of input-aware attention}: We first remove the input-aware personal aggregator and summarize the personal embedding by averaging the representation of each historical behavior. As shown in Figure~\ref{fig:abl}, this model performs worse than the full \ours{}, indicating that the input-aware aggregator can better capture user patterns according to the current input. Nevertheless, it is worth noting that even without this component, \ours{} still achieves strong results compared to baselines, suggesting that the user's overall behavior patterns are crucial for personalized language generation.

(2) \textbf{Impact of instruction embedding}: Next, we remove the instruction embedding $[\bI]$ from the LLM input in Equation~(\ref{eq:loss}). Intriguingly, this variant also outperforms baselines, indicating that the primary improvements of the \ours{} stem from the personal embedding from user histories.
However, the performance decline highlights that the instruction embedding helps the model disentangle global task-related knowledge from user-specific patterns, thereby enhancing personalization performance.

\begin{table}[!t]
 \center
 \small
 \caption{Performance of \ours{} integrated with retrieval on the validation set.} 
 \label{tlb:integration}
  \begin{tabular}{lccc}
  	\toprule
        {Dataset} & {Metric} & {\ours{}} & {\ours{} + Retrieval} \\
        \midrule
        LaMP-1 & Accuracy $\uparrow$ & 0.680 & \textbf{0.687} \\
        \midrule
        \multirow{2}{*}{LaMP-2} & Accuracy $\uparrow$ & \textbf{0.565} & 0.545\\
         & F1 $\uparrow$ & \textbf{0.501} & 0.485 \\
        \midrule
        \multirow{2}{*}{LaMP-3} & MAE $\downarrow$ & 0.231 & \textbf{0.215} \\
         & RMSE $\downarrow$ & 0.534 & \textbf{0.506}\\
        \midrule
        \multirow{2}{*}{LaMP-4} & ROUGE-1 $\uparrow$ & 0.216 & \textbf{0.220} \\
         & ROUGE-L $\uparrow$ & 0.197 & \textbf{0.203} \\
        \midrule
        \multirow{2}{*}{LaMP-5} & ROUGE-1 $\uparrow$ & 0.487 & \textbf{0.498} \\
         & ROUGE-L $\uparrow$ & 0.436 & \textbf{0.448} \\
        \midrule
        \multirow{2}{*}{LaMP-7} & ROUGE-1 $\uparrow$ & 0.534 & \textbf{0.547}\\
         & ROUGE-L $\uparrow$ & 0.484 & \textbf{0.495} \\
	\bottomrule
  \end{tabular}
\end{table}

\paragraph{Integration with Retrieval-based Strategy}
\label{para:retrieval}
In our experiments, we observe that the retrieval-based personalization methods can yield improvements over non-personalization methods. Therefore, we investigate whether integrating our \ours{} method with retrieval-based strategies can further enhance performance. Specifically, we first use the BGE-base-en-v1.5 model~\cite{bge_1, bge_2} to retrieve the most relevant historical behavior from the user's history based on the current input. Then, the retrieved content $h^u_k$ is appended to the input as demonstrations for the LLM to reference for producing personalized outputs. In this manner, the inputs can be formatted as $\bX^u_i = [\textrm{Emb}_{\textrm{LLM}}(h^u_{k});\bI;\bP^u;\textrm{Emb}_{\textrm{LLM}}(x^u)]$. We train the model using the same training data and refer to this model as ``\ours{} + Retrieval''. The results are shown in Table~\ref{tlb:integration}. 
Overall, the integration of retrieval-based strategies with the \ours{} model leads to further performance gains over the original \ours{} method. Indeed, \ours{} provides a coarse-grained user style embedding, capturing general user habits and preferences. In contrast, retrieval-based methods offer fine-grained, task-specific historical contexts that help retrieve knowledge relevant to the current task. Therefore, combining these approaches allows for more effective personalized generation. This raises a new research question of how to optimize the use of coarse-grained user embeddings versus fine-grained retrieved references, suggesting a direction for future research.

\begin{table}
 \small
 \centering
 \caption{Performance of \ours{} with history selection  on the validation set.}
 \label{tlb:select}
  \begin{tabular}{lccc}
  	\toprule
        {Dataset} & {Metric} & {\ours{}} & {\ours{} with selection} \\
        \midrule
        LaMP-1 & Accuracy $\uparrow$ & \textbf{0.680} & 0.675 \\
        \midrule
        \multirow{2}{*}{LaMP-2} & Accuracy $\uparrow$ & \textbf{0.565} & 0.492\\
         & F1 $\uparrow$ & \textbf{0.501} & 0.441 \\
        \midrule
        \multirow{2}{*}{LaMP-3} & MAE $\downarrow$ & \textbf{0.231} & 0.239 \\
         & RMSE $\downarrow$ & \textbf{0.534} & 0.542\\
        \midrule
        \multirow{2}{*}{LaMP-4} & ROUGE-1 $\uparrow$ & \textbf{0.216} & 0.205 \\
         & ROUGE-L $\uparrow$ & \textbf{0.197} & 0.188 \\
        \midrule
        \multirow{2}{*}{LaMP-5} & ROUGE-1 $\uparrow$ & \textbf{0.487} & 0.485 \\
         & ROUGE-L $\uparrow$ & \textbf{0.436} & \textbf{0.436} \\
        \midrule
        \multirow{2}{*}{LaMP-7} & ROUGE-1 $\uparrow$ & \textbf{0.534} & 0.530 \\
         & ROUGE-L $\uparrow$ & \textbf{0.484} &  0.477 \\
	\bottomrule
  \end{tabular}
\end{table}

\paragraph{History Selection Study}
\label{para:sel}
As discussed in Section~\ref{sec:intro}, retrieval-based personalization approaches only select historical content most relevant to the current task input to serve as demonstrations for the LLM, which may hinder the model from capturing the user's broader interests. To explore this, we modify our input-aware personal aggregator module to select only the top-4 history embeddings, $\bh^u_i$, based on their associated weights, $w_i$, to construct the personal embedding $\bP^u$. This setting is consistent with retrieval-based personalized LLMs, which rely on a small set of top historical behaviors. We refer to this variant as 
 ``\ours{} with Selection''. The results are presented in Table~\ref{tlb:select}.
We can observe that the performance of \ours{} model decreases when using only the top-4 history embeddings to build the user’s personal embedding. This suggests that the selective usage of histories can impair the model’s ability to capture general user patterns, leading to sub-optimal outputs. In contrast, aggregating all histories, as in the original \ours{} model, provides a more comprehensive representation of the user’s preferences, resulting in improved performance. We further conduct experiments to analyze the impact of the number of selected histories, which can be found in Appendix~\ref{apd:selexp}.

\begin{table}
 \small
 \centering
 \caption{Performance of \ours{} with users with different history length.}
 \label{tlb:his_len}
  \begin{tabular}{lcc}
  	\toprule
        {User Group} & {LaMP-3 RMSE $\downarrow$} & {LaMP-5 ROUGE-1 $\uparrow$} \\
        \midrule
        Short & 0.548 & 0.474 \\
        Medium & 0.531 & 0.482 \\
        Long & \textbf{0.522} & \textbf{0.499} \\
        \midrule
        Overall & 0.534 & 0.487 \\
	\bottomrule
  \end{tabular}
\end{table}

\paragraph{History Length Impact Study}
In this part, we investigate the impact of the amount of user historical data on model performance. Specifically, we group users in LaMP-3 Personalized Product Rating Task (a classification task) and LaMP-5 Personalized Scholarly Title Generation Task (a generation task) into three groups (short/medium/long) according to their history length by 1:1:1 ratio. We show the performances on each group on the validation set in Table~\ref{tlb:his_len}. We can observe that the performance of our PPlug model increases with the length of user histories because longer histories can provide more information about user interests. Besides, the performance of the model on users with different histories is generally robust, suggesting that \ours{} is also effective even when users have limited histories.

\section{Conclusion}
\label{sec:conclusion}
In this work, we propose a persona-plug (\ours{}) model for personalized language generation. In \ours{} model, we devise a lightweight and plug-and-play user embedder module to encode a user's all historical behaviors to dense vectors and then aggregate them into one single user personal embedding in an input-aware manner. We believe this distinct personal embedding for each user can represent their general linguistic styles and habits in all histories and guide LLMs to personalize their outputs. Experimental results on the LaMP benchmark show that the proposed model can significantly outperform existing retrieval-based LLM models. 

\section*{Acknowledgement}
Yutao Zhu and Zhicheng Dou are the corresponding authors. This work was supported by Beijing Municipal Science and Technology Project No. Z231100010323009, National Natural Science Foundation of China No. 62272467, Beijing Natural Science Foundation No. L233008, and the fund for building world-class universities (disciplines) of Renmin University of China. The work was partially done at the Beijing Key Laboratory of Research on Large Models and Intelligent Governance.

\section*{Limitations}
In this study, we propose a novel personalized LLM model that encodes a specific user's all history into user-specific personal embeddings and attaches it to inputs for LLMs to perform personalization. We admire several limitations in this work for further exploration and investigation.

First, in our \ours{} model, we only represent histories at the behavior level. However, some terms and phrases that users frequently use in their histories can also help us to capture general user patterns and styles. A potential future work is to augment the personal embedding with fine-grained term-level information. Second, as we experimented and discussed in Section~\ref{para:retrieval}, \ours{} can be integrated with retrieval-based methods to improve performance. In the future, we can study when to utilize the user embedding and when to use the in-context retrieved references for personalizing LLM-generated outputs.

\section*{Ethical Considerations}
The LaMP benchmark used in our experiments is publicly available on the Web and does not have privacy concerns. For the applications of personalized language generation, they usually require the collection of user historical data, which may cause privacy leakage problems. Although there may exist risks of abusing and leaking user data in personalization tasks, our proposed \ours{} model indeed alleviates or even solves the problems. LLM service providers only need to release the tuned user embedder model to users, and users can build and upload their specific personal embeddings by themselves to guide LLMs in providing personalized results. During this process, users do not need to upload their own historical text data. In contrast, previous personalized LLM approaches need to obtain user data for retrieval or tuning.

\bibliography{custom}

\appendix

\begin{table*}[!ht]
 \center
 \small
 \caption{Data statistics of the six experimented tasks in the LaMP benchmark.} 
 \label{tlb:sta}
 \setlength{\tabcolsep}{1.5mm}{
  \resizebox{0.99\linewidth}{!}{\begin{tabular}{llrrrrrrr}
  	\toprule
        {\textbf{Task}} & {\textbf{Task Type}} & \textbf{\#Train} & \textbf{\#Validation} & 
        \textbf{\#Test} &
        \textbf{Input Length} & \textbf{Output Length} & \textbf{History Length} & \textbf{\#Classes} \\
        \midrule
        LaMP-1 & Binary classification & 6,542 & 1,500 & 1,500 & 51.43 $\pm$ 5.70 & - & 84.15 $\pm$ 47.54 & 2 \\
        LaMP-2 & Categorical classification & 5,073 & 1,410 & 1,557 & 92.39 $\pm$ 21.95 & - & 86.76 $\pm$ 189.52 & 15 \\
        LaMP-3 & Ordinal classification & 20,000 & 2,500 & 2,500 & 128.18 $\pm$ 146.25 & - & 185.40 $\pm$ 129.30 & 5 \\
        LaMP-4 & Text generation & 12,500 & 1,500 & 1,800 & 29.97 $\pm$ 12.09 &10.07 $\pm$ 3.10 & 204.59 $\pm$ 250.75 & - \\
        LaMP-5 & Text generation & 14,682 & 1,500 & 1,500 & 162.34 $\pm$ 65.62 & 9.71 $\pm$ 3.21 & 87.88 $\pm$ 53.63 & - \\    
        LaMP-7 & Text generation & 13,437 & 1,498 & 1,500 & 29.72 $\pm$ 7.01 & 16.96 $\pm$ 5.67 & 15.71 $\pm$ 14.85 & - \\
 \bottomrule
  \end{tabular}}
}
\end{table*}

\section{Dataset Details}
Detailed statistics for the six tasks are provided in Table~\ref{tlb:sta}. The formats of input, output, and user histories of the six tasks are shown in Table~\ref{tlb:format}.

\label{apd:dataset}
\begin{table*}[!hpt]
 \center
 \small
 \caption{Format of the input, output, and user histories of six tasks in the LaMP benchmark. \textit{Italic text} will be
replaced with realistic data for each task during training and inference. } 
 \label{tlb:format}
  \begin{tabular}{llll}
  	\toprule
        {\textbf{Task}} & {\textbf{Input}} & {\textbf{Output}} & {\textbf{User History}} \\
        \midrule
        \makecell[l]{LaMP-1} & \makecell[l]{For an author who has written the paper with the title \\ ``\textit{\{title\}}'', which reference is related? Just answer \\ with [1] or [2] without explanation. \\ {[1]}: ``\textit{\{reference1\}}'' [2]: ``\textit{\{reference2\}}''} & \makecell[l]{[1]} & \makecell[l]{title: \textit{\{title\}} \\ abstract: \textit{\{abstract\}}} \\ 
        \midrule
        \makecell[l]{LaMP-2} & \makecell[l]{Which tag does this movie relate to among the following tags? \\ Just answer with the tag name without further explanation.\\ tags: [sci-fi, based on a book, comedy, action, twist ending, \\ dystopia, ...] description: \textit{\{movie\}}} & \makecell[l]{sci-fi} & \makecell[l]{description: \textit{\{movie\}} \\ tag: \textit{\{tag\}}} \\ 
        \midrule
        \makecell[l]{LaMP-3} & \makecell[l]{What is the score of the following review on a scale of 1 to 5? \\ Just answer with 1, 2, 3, 4, or 5 without further explanation. \\ review: \textit{\{review\}}} & \makecell[l]{3} & \makecell[l]{text: \textit{\{review\}} \\ score: \textit{\{score\}}} \\ 
        \midrule
        \makecell[l]{LaMP-4} & \makecell[l]{Generate a headline for the following article: \textit{\{article\}}} & \makecell[l]{How I \\ Got 'Rich'} & \makecell[l]{title: \textit{\{title\}} \\ text: \textit{\{article\}}} \\ 
        \midrule
        \makecell[l]{LaMP-5} & \makecell[l]{Generate a title for the following \\ abstract of a paper: \textit{\{abstract\}}} & \makecell[l]{Distributed \\Partial Clustering} & \makecell[l]{title: \textit{\{title\}} \\ text: \textit{\{abstract\}}} \\ 
        \midrule
        \makecell[l]{LaMP-7} & \makecell[l]{Paraphrase the following tweet without any explanation \\ before or after it: \textit{\{tweet\}}} & \makecell[l]{gotta make the \\ most of my last \\ full day in ktown} & \makecell[l]{text: \textit{\{tweet\}}} \\ 
 \bottomrule
  \end{tabular}
\end{table*}

\section{More Experimental Results}
\label{apd:exp}

\subsection{Complete Ablation Study}
\label{apd:ablexp}
We show the complete ablation results on the LaMP benchmark in Table~\ref{tlb:complete_abl}. Our \ours{} model generally outperforms all ablation models. The results are consistent with the results in Section~\ref{para:abl}.

\begin{table*}[!ht]
 \centering
 \small
 \caption{Overall performance of ablation models on the validation set. We use ``Acc'' to denote Accuracy and ``R'' to denote ROUGE, respectively.
 }\label{tlb:complete_abl}  \setlength{\tabcolsep}{1.2mm}{
  \begin{tabular}{lccccccccccccc}
  	\toprule
        \multirow{2}{*}[-2pt]{Model} & LaMP-1 & \multicolumn{2}{c}{LaMP-2} & \multicolumn{2}{c}{LaMP-3} & \multicolumn{2}{c}{LaMP-4} & \multicolumn{2}{c}{LaMP-5} & \multicolumn{2}{c}{LaMP-7} & \multirow{2}{*}[-2pt]{\# Best} \\
        \cmidrule(lr){2-2} \cmidrule(lr){3-4} \cmidrule(lr){5-6} \cmidrule(lr){7-8} \cmidrule(lr){9-10} \cmidrule(lr){11-12}
        & Acc $\uparrow$ &  Acc $\uparrow$ &  F1 $\uparrow$ &  MAE $\downarrow$ &  RMSE $\downarrow$  & R-1 $\uparrow$ & R-L $\uparrow$ & R-1 $\uparrow$ & R-L $\uparrow$ & R-1 $\uparrow$ & R-L $\uparrow$ \\
        \midrule
        \ours{} & \textbf{0.6800} & \textbf{0.5652} & \textbf{0.5014} & 0.2312 & 0.5337 & \textbf{0.2162} & \textbf{0.1972} & \textbf{0.4869} & 0.4359 & 0.5338 & \textbf{0.4836} & 7\\
        \quad \textit{w/o.} IE & 0.6786 & 0.5510 & 0.4834 & \textbf{0.2304} & \textbf{0.5238} & 0.2142 & 0.1960 & 0.4852 & 0.4350 & 0.5301 & 0.4781 & 2\\
        \quad \textit{w/o.} IA & 0.6786 & 0.5644 & 0.4924 & 0.2320 & 0.5333 & 0.2160 & 0.1961 & 0.4852 & \textbf{0.4363} & \textbf{0.5351} & 0.4818 & 2\\    
	\bottomrule
  \end{tabular}}
\end{table*}

\subsection{Further History Selection Study}
\label{apd:selexp}

\begin{figure}
    \centering
    \includegraphics[width=\linewidth]{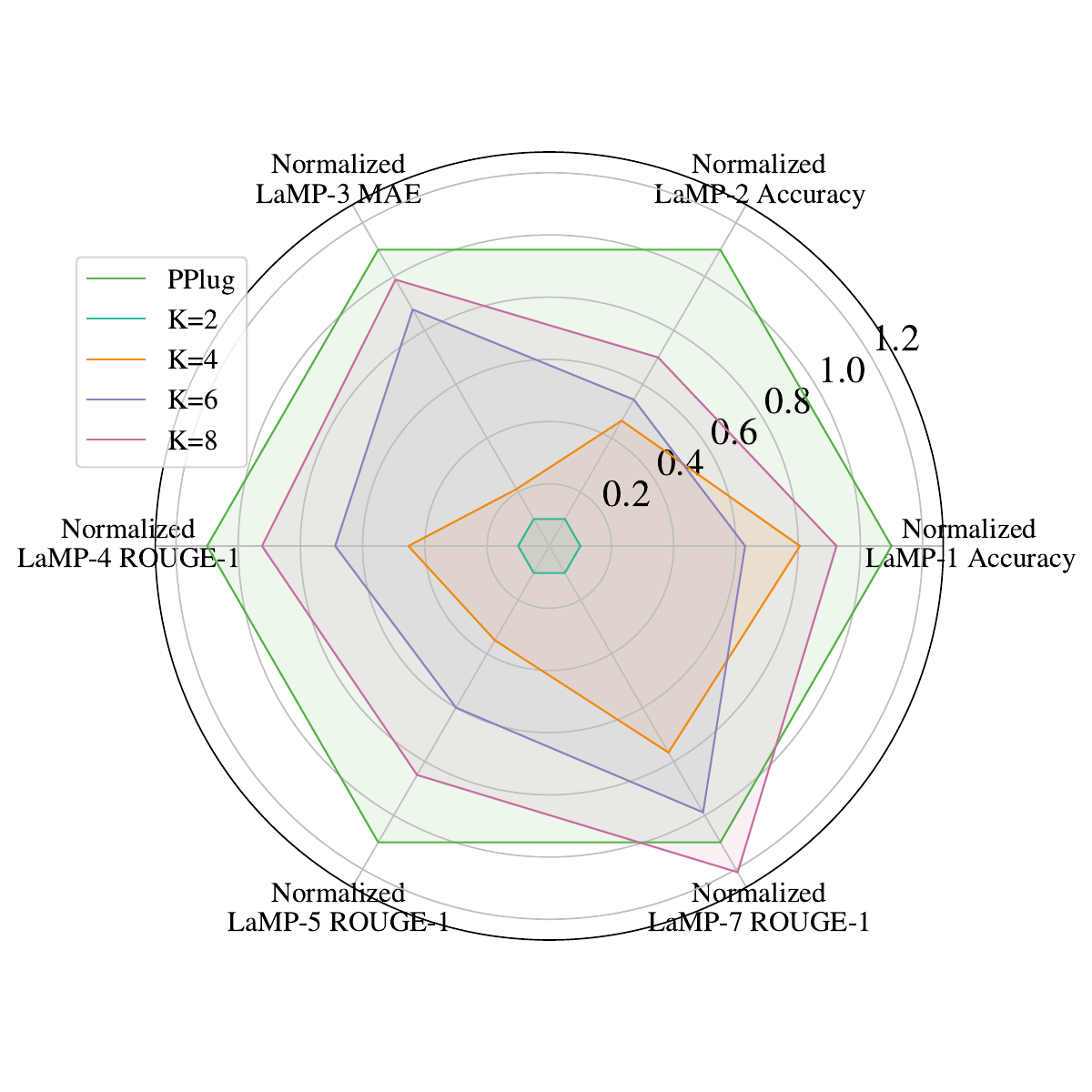}
    \caption{Performance of \ours{} selecting only top-$K$ user histories on the validation set.}
    \label{fig:selection}
\end{figure}

In this section, we further analyze the impact of the number of histories used to build the personal embedding in Section~\ref{para:sel}. Specifically, we modify our input-aware personal aggregator module to utilize only the top-$K$ history embeddings for constructing personal embedding, where $K$ ranges from 2 to 8. For convenient comparison, we normalize the results $R$ on each task by: 
\begin{equation}
\begin{aligned}
    R_{\textrm{normalize}} = \frac{R-R_{K=2}}{R_{\textrm{all}}-R_{K=2}} + \epsilon
\end{aligned}
\end{equation}
where $R_{\textrm{all}}$ denotes the result of \ours{} model using all histories,$R_{K=2}$ denotes the result of using only top-2 histories. We set $\epsilon=0.1$. The results are shown in Figure~\ref{fig:selection}.

We can observe that with the number of utilized histories increasing, the performance of the \ours{} model keeps rising. However, the performance is consistently lower than \ours{} model using all histories except the LaMP-7 Personalized Tweet Paraphrasing task. The reason may be that the user history length is shorter compared with other tasks, thus the selection manipulation may not break the overall user patterns severely but function as a denoising operation.

\end{document}